\documentclass[letterpaper, 10 pt, conference]{ieeeconf}  % Comment this line out if you need a4paper
\IEEEoverridecommandlockouts                              % This command is only needed if 
\usepackage{graphics} % for pdf, bitmapped graphics files
\usepackage{epsfig} % for postscript graphics files

\usepackage{stfloats}
\usepackage{times} % assumes new font selection scheme installed
\usepackage{amsmath} % assumes amsmath package installed
\usepackage{amssymb}  % assumes amsmath package installed
\usepackage[hang]{subfigure} % two figures horizontally
\usepackage{cite}
\usepackage{siunitx}
\usepackage{soul}
\usepackage{color}
\usepackage{xcolor}
\usepackage{mathrsfs}
\usepackage[ruled,linesnumbered]{algorithm2e}
\usepackage[font=scriptsize]{caption}

\soulregister\ref7  % 针对\ref命令
\soulregister\cite7 
\soulregister\eqref7
\soulregister\prettyref7

\usepackage{multirow}
\usepackage{algpseudocode}
\usepackage{makecell}
\usepackage{booktabs}
\usepackage{threeparttable}
\usepackage{hyperref}
\usepackage{endnotes}

\newcommand{\eg}{\textit{e}.\textit{g}.}

\newcommand{\ie}{\textit{i}.\textit{e}.}

\bstctlcite{IEEEexample:BSTcontrol}

\usepackage{prettyref}
\newrefformat{fig}{Fig.~\ref{#1}}
\newrefformat{sec}{Sec.~\ref{#1}}
\newrefformat{tab}{Tab.~\ref{#1}}
\newrefformat{algo}{Algo.~\ref{#1}}
\newrefformat{eq}{(\ref{#1})}

\usepackage{color}

\title{\LARGE \bf
\textcolor{violet}{D}istillation-\textcolor{violet}{PPO}: A Novel Two-Stage Reinforcement Learning Framework for Humanoid Robot Perceptive Locomotion}

\author{Qiang Zhang$^{1,2,\ast}$,  Gang Han$^{1,\ast,\clubsuit}$, Jingkai Sun$^{1,2,\ast}$, Wen Zhao$^{1,\ast}$, \\Chenghao Sun$^{1}$, Jiahang Cao$^{2}$, Jiaxu Wang$^{2}$, Yijie Guo$^{1,\dagger}$, Renjing Xu$^{2,\dagger}$}

\begin{document}
\maketitle

\footnotetext[1]{The authors are with Beijing Innovation Center of Humanoid Robotics Co. Ltd. {\tt\small Jony.Zhang@xhumanoid.com}}
\footnotetext[2]{The authors are with The Hong Kong University of Science and Technology (Guangzhou), China. {$^{\ast}$ is corresponding authors. $^{\dagger}$ are the corresponding authors. $^{\clubsuit}$ is the project leader.} {\tt\small  \ qzhang749@connect.hkust-gz.edu.cn}}

\thispagestyle{empty}
\pagestyle{empty}

%%%%%%%%%%%%%%%%%%%%%%%%%%%%%%%%%%%%%%%%%%%%%%[section]%%%%%%%%%%%%%%%%%%%%%%%%%%%%%%%%%%%%%%%%%%%%%%%%%%%%%%%%%%
\begin{abstract}

In recent years, humanoid robots have garnered significant attention from both academia and industry due to their high adaptability to environments and human-like characteristics. With the rapid advancement of reinforcement learning, substantial progress has been made in the walking control of humanoid robots. However, existing methods still face challenges when dealing with complex environments and irregular terrains.
In the field of perceptive locomotion, existing approaches are generally divided into two-stage methods and end-to-end methods.
Two-stage methods first train a teacher policy in a simulated environment and then use distillation techniques, such as DAgger, to transfer the privileged information learned as latent features or actions to the student policy. End-to-end methods, on the other hand, forgo the learning of privileged information and directly learn policies from a partially observable Markov decision process (POMDP) through reinforcement learning. 
However, due to the lack of supervision from a teacher policy, end-to-end methods often face difficulties in training and exhibit unstable performance in real-world applications.
This paper proposes an innovative two-stage perceptive locomotion framework that combines the advantages of teacher policies learned in a fully observable Markov decision process (MDP) to regularize and supervise the student policy. At the same time, it leverages the characteristics of reinforcement learning to ensure that the student policy can continue to learn in a POMDP, thereby enhancing the model's upper bound.
Our experimental results demonstrate that our two-stage training framework achieves higher training efficiency and stability in simulated environments, while also exhibiting better robustness and generalization capabilities in real-world applications.

\end{abstract}

%%%%%%%%%%%%%%%%%%%%%%%%%%%%%%%%%%%%%%%%%%%%%%[section]%%%%%%%%%%%%%%%%%%%%%%%%%%%%%%%%%%%%%%%%%%%%%%%%%%%%%%%%%%
\section{Introduction}
\label{sec:intro}

In recent years, humanoid robots have garnered significant attention due to their human-like form and high adaptability to human living and working environments. The development of reinforcement learning and deep learning has brought revolutionary changes to the walking control of humanoid robots, enabling them to go beyond performing pre-set actions and tasks, and offering more possibilities. Among the tasks for humanoid robots, locomotion is one of the most critical and challenging. It requires robots to adapt to complex and dynamic environments through their own actions, completing tasks such as walking, running, and jumping.

\begin{figure}[t]
	\centering
	\includegraphics[width=\linewidth]{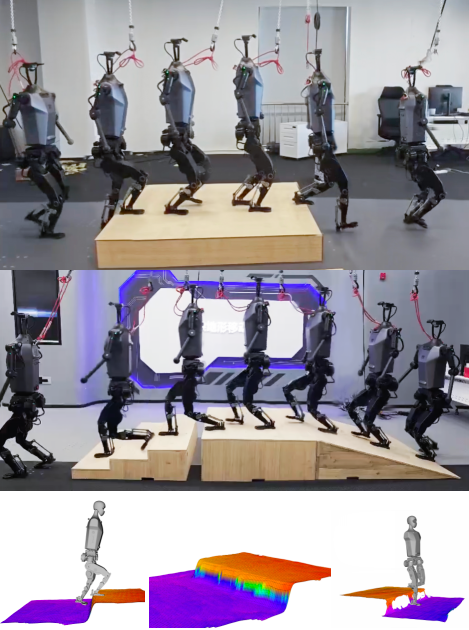}
 \caption{We demonstrate the walking capabilities of the humanoid robot Tien Kung on different terrains after being trained with ~\textcolor{violet}{D-PPO} in the figure. The top part of the figure shows Tien Kung navigating a platform nearly as high as its shins. The middle part illustrates Tien Kung's ability to walk on a sloped surface and cross a small ditch. The bottom part displays the terrain reconstruction by our perception system and the robot's posture while overcoming obstacles. It is clear that Tien Kung's posture varies across different terrains. On flat ground, Tien Kung maintains a relatively straight-knee posture to achieve a broader field of view. When crossing obstacles, Tien Kung bends its knees to reduce impact and swings its arms to maintain balance.}
	\label{fig:teaser}
	% \vspace{-0.55cm}
\end{figure}

The locomotion task of humanoid robots can be categorized as bipedal walking, which falls under the broader category of legged robot locomotion~\cite{xie2023learning,li2024reinforcement,zhang2024whole,siekmann2021sim,gong2019feedback}. In the traditional field of optimal control, locomotion tasks are typically modeled as optimal control problems~\cite{khazoom2024tailoring}. This involves establishing dynamic models and optimization objective functions, and using nonlinear programming methods to solve for the optimal joint torques or joint angles. However, this approach often requires extensive domain knowledge and manually designed features, making it difficult to adapt to complex and dynamic environments.
The advancement of reinforcement learning has introduced new approaches to solving locomotion tasks. By training policy networks in simulated environments and then transferring them to real robots, autonomous walking of humanoid robots can be achieved. For perceptive locomotion based on neural networks, existing solutions can be broadly divided into two categories: two-stage methods~\cite{lee2020learning,kumar2021rma,cheng2024extreme,hoeller2024anymal,jenelten2024dtc} and end-to-end methods~\cite{yang2023neural}.

Two-stage methods first train a teacher policy in a simulated environment and then use distillation techniques to transfer the privileged information learned as latent features or actions to the student policy. The advantage of this approach is that the strategy advantages learned by the teacher policy in a fully observable Markov decision process can regularize and supervise the student policy, thereby improving its performance. However, this method also has some drawbacks. Firstly, the teacher policy may limit the potential of the student policy, making it only asymptotically approach the behavior of the teacher policy without surpassing it. This contradicts the core idea of reinforcement learning, which is to learn better policies through continuous trial and error interactions with the environment. Additionally, the teacher policy usually requires a large amount of privileged information, most of which can only be obtained in simulations. The inaccuracies of the simulation engine in representing the real world may lead to erroneous guidance from the teacher policy. Most critically, the teacher policy is not infallible; when encountering unseen scenarios, it may fail to provide correct guidance due to a lack of sufficient prior knowledge. In such cases, the student policy has no room for adjustment and can only follow the teacher policy's guidance, resulting in poor performance in real-world scenarios.

End-to-end methods forgo the learning of privileged information and directly learn policies from a partially observable Markov decision process (POMDP) through reinforcement learning. The advantage of this approach is that it avoids the limitations imposed by the teacher policy, allowing the student policy to freely explore and learn. However, this method also has some drawbacks. Firstly, learning policies from a POMDP is challenging because the agent cannot obtain complete environmental information, making the learning process more difficult and unstable. Additionally, due to the lack of supervision from privileged information, the student policy may struggle to learn high-quality policies, resulting in performance that is inferior to that of two-stage methods.

Both two-stage and end-to-end methods have their own advantages and disadvantages. A natural question arises: is it possible to combine the strengths of both approaches to enhance the locomotion capabilities of humanoid robots? This is precisely the goal of our paper. We believe that in the two-stage method, the supervision signal from the teacher policy serves as an excellent regularization signal, which can regularize and guide the student policy during training, ensuring that the student policy converges in the correct direction. At the same time, during the learning process of the student policy, we fully leverage the characteristics of reinforcement learning to ensure that the student policy can continue to learn in a partially observable Markov decision process, thereby improving the model's upper bound.

We propose a two-stage training framework named Distillation-PPO (D-PPO). In the first stage, we train a teacher policy in a fully observable Markov decision process and use distillation methods to convert privileged information into latent features or actions to supervise the student policy's learning. In the second stage, we train the student policy in a partially observable Markov decision process, combining the supervision signals from the teacher policy with reinforcement learning rewards to further enhance the student policy's performance.
Our experimental results demonstrate that D-PPO achieves higher training efficiency and stability in simulated environments, while also exhibiting better robustness and generalization capabilities on actual robots.

We demonstrate humanoid robot Tien Kung's walking capabilities on different terrains after being trained with D-PPO in Fig.~\ref{fig:teaser}. In summary, the contributions of this paper can be summarized as follows:

\begin{enumerate}
    \item We propose~\textcolor{violet}{Distillation-PPO (D-PPO)}, a two-stage training framework, that combines the supervision signals from the teacher policy with the reinforcement learning reward mechanism. 
    \item We conducted numerous valuable real-world experiments with humanoid robots walking on various terrains. Through extensive experiments, we demonstrate that D-PPO achieves higher training efficiency and stability in both training and simulated environments. Moreover, it shows stronger robustness and generalization capabilities in real-world robot applications.
\end{enumerate}

This paper not only introduces a new training method but also showcases its superior performance in complex environments through practical experiments, providing new insights and methods for future humanoid robot locomotion control. Additionally, our method can be easily extended and transferred to other legged robots.

%%%%%%%%%%%%%%%%%%%%%%%%%%%%%%%%%%%%%%%%%%%%%%[section]%%%%%%%%%%%%%%%%%%%%%%%%%%%%%%%%%%%%%%%%%%%%%%%%%%%%%%%%%%

\begin{figure*}[tbp]
	\centering
	\includegraphics[width=.8\linewidth]{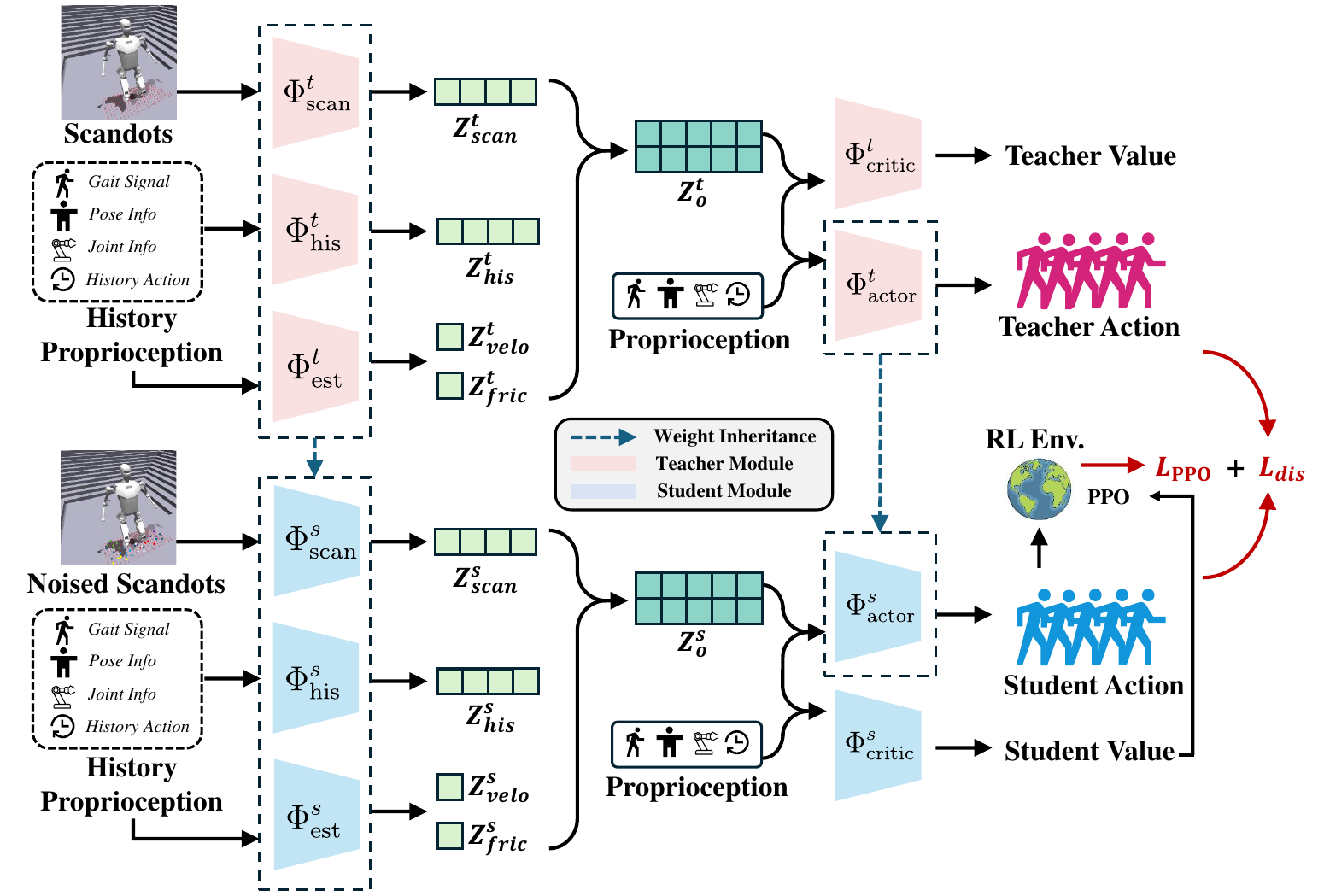}
 \caption{The training framework of ~\textcolor{violet}{Distillation-PPO} adopts a symmetric structure for both the teacher and student networks. We did not introduce excessive privileged information into the teacher network because we found that much of the privileged information in simulations is inaccurate and can limit the performance of the student network on actual robots. We extensively use historical information to estimate the state and fully inherit the structure and parameters of the teacher network for initialization, similar to a self-distillation structure. During training, we simulate real-world conditions by increasing the noise in the input information of the student network (domain randomization). In our framework, the distillation loss acts as a regularization term, ensuring that the student network does not deviate during training, while the reinforcement learning loss acts as a reward, ensuring the exploration efficiency and performance upper bound of the student network during training.}
	\label{fig:main}
	\vspace{-0.55cm}
\end{figure*}

\section{Related Work}

\subsection{Reinforcement Learning on Legged Robots}
With the advent of Reinforcement Learning (RL) and Imitation Learning (IL), new methods of automating the learning process for humanoid locomotion have emerged. The utilization of multiple reward functions in RL has shown promise.  \cite{jeon2023benchmarking} studies the RL reward shaping methods which are robust to scaling on the high-dimensional system (\eg, humanoid robot). This enhancement simplifies and accelerates the tuning process of reward weights. However, this approach, which relies only on the reward function, has the potential to generate unnatural motions.  By introducing reference trajectories~\cite{siekmann2020learning} and periodic rewards~\cite{siekmann2021blind,siekmann2021sim}, these methods allow for more natural locomotion of bipedal robots without the need for complex modeling or reward function design. However, such periodic rewards can only be used for the control of bipedal robots and cannot be extended to humanoid robots, \ie, locomotion that involves whole-body control. ~\cite{guo2023decentralized} proposes a novel decentralized framework inspired by biology to control different parts of humanoid robots separately. This work shows that the decentralized control policy can be more robust to local motor malfunctions and easy to transfer to new tasks. ~\cite{li2023robust} further elucidates the capabilities of bipedal robots, demonstrating proficiency in executing tasks that require jumping.
In contrast to humanoid robots, quadrupedal robots exhibit a simplistic design that inherently facilitates easier stabilization. RL-based approaches allow complex tasks to be performed in a variety of terrains and environments, such as parkour~\cite{zhuang2023robot,cheng2023extreme,hoeller2023anymal} and mountaineering~\cite{lee2020learning}.

\subsection{Perceptive Locomotion on Legged Robots}
Perception is a key factor in stable and precise locomotion. Compared with blind methods, perceptive locomotion can enable the robots to make decisions in advance rather than through contact between the feet and the terrain. Avoid being disturbed by contact, the control algorithm can achieve perfect performance in task space such as velocity tracking~\cite{cheng2024extreme} or navigation~\cite{hoeller2024anymal,zhang2024resilient}. \cite{imai2022vision,yang2021learning} apply depth images in end-to-end quadrupedal locomotion frameworks, which utilize Transformer or other networks to fuse perceptive information and proprioception. However, these methods are usually limited by depth image sample efficiency and sim-to-real gap between images in simulation and the world. \cite{hoeller2024anymal, hoeller2022neural} propose an additional network to reconstruct terrain point cloud from raw lidar data. Subsequently, the navigation and control modules receive the reconstruction results as observation. \cite{zhuang2023robot,cheng2024extreme} propose two-stage distillation-based methods to convert depth images to scan dots or visual constraints and apply multiple data augmentation to reduce the sim-to-real gaps. \cite{duan2024learning} combines a blind policy and a vision-based modulator. The blind policy outputs basic action and collects data to train a predictor to convert the depth images to scan dots. The vision-based modulator will produce modulating action of the desired motor position as well as clock actions to modify the clock parameters. However, distillation-based methods only use supervised loss to force the student to execute the same actions but hard to tune the performance via rewards. The data collected by blind policy exist the domain gap between networks with vision-based modulator, which reduces the performance of these methods like ~\cite{duan2024learning}.
%%%%%%%%%%%%%%%%%%%%%%%%%%%%%%%%%%%%%%%%%%%%%%[section]%%%%%%%%%%%%%%%%%%%%%%%%%%%%%%%%%%%%%%%%%%%%%%%%%%%%%%%%%%
\section{Background}

\subsection{Reinforcement Learning on Robots}
We model humanoid locomotion as a Markov Decision Process (MDP) characterized by $(\mathcal{S}, \mathcal{A}, \mathcal{R}, p, \gamma)$. $\mathcal{S}$ represents the state space, $\mathcal{A}$ is the action space, and $\mathcal{R}$ is the reward function. The transition probabilities from the current state $s_t$ to the next state $s_{t+1}$ are denoted by $p$, while $\gamma \in [0,1]$ is the discount factor for future rewards. At each time step $t$, the policy selects an action based on the current state, resulting in a transition to the next state $s_{t+1}$ according to the transition dynamics $s_{t+1} \sim p(s_{t+1}|s_t,a_t)$. The goal of training is to optimize the policy parameters $\theta$ to maximize the cumulative reward: 
\begin{equation} 
\textnormal{arg}\max_{\theta} \mathbb{E}{(s_t,a_t) \sim p_\theta(s_t,a_t)} \left[ \sum_{t=0}^{T-1} \gamma^t r_t\right] 
\end{equation}
In humanoid locomotion scenarios, actions are defined by the desired position of each actuated joint. Observations include proprioception about humanoid robots and the perception of equipping the robots with the ability to deal with complex terrain.

\subsection{Elevation Map and Scan Dots}
An occupancy grid is commonly used for navigation on flat surfaces with obstacles. This 2D map indicates whether each cell is occupied, enabling robots to plan paths to target points while avoiding obstacles. However, this approach is inadequate for environments with uneven terrain, such as slopes or steps~\cite {miki2022elevation}. To address this limitation, a 2.5D elevation map is often employed, which records the ground height for each cell. In the elevation map, a 2D grid captures height data to represent the geometry of the terrain. To further enhance this representation of complex structures like bridges. a multilevel surface map is introduced. Compared with the 3D grid that can model more complex structures than an elevation map, it also demands significantly more memory for the same grid resolution.

To feed the information of terrain into policy network, we compress the elevation map into scan dots. By sampling points at one-meter intervals on the elevation map, we obtain scan dots vector $m\in\mathbb{R}^{441}$, which represents the terrain height grid map centered around our robot.

%%%%%%%%%%%%%%%%%%%%%%%%%%%%%%%%%%%%%%%%%%%%%%[section]%%%%%%%%%%%%%%%%%%%%%%%%%%%%%%%%%%%%%%%%%%%%%%%%%%%%%%%%%%
\section{Method}

\subsection{Two-Stage Reinforcement Learning with Teacher Distillation}
\textbf{Teacher training stage}. In the initial stage, we directly obtain scan dots from the simulator and use Conv1D to compress the scan vectors into a 32-dimensional latent space. In addition, we apply an external encoder that compresses the 50-frame history state into a 32-dimensional latent space. We then use a combined MLP to process both the state history latent and perception latent. At this stage, the map is clean and precise, enabling the robot to navigate complex terrain geometries effectively. Simultaneously, we follow the approach described in our prior work~\cite{zhang2024whole} to construct a periodic system. However, using fixed periodic parameters can lead to conflicts between navigating challenging terrain and maintaining stable locomotion. For instance, when both the desired velocity command and the periodic frequencies are fixed, the desired foot distance in each step becomes predetermined. In such scenarios, even if the policy perceives dangerous geometries, such as the edge of a step, the robot may still make contact with it.

\textbf{Student training stage}.
In the next phase, the student policy is required to process noisy perception results to bridge the gap between simulation and the real world. However, the noise level significantly impacts policy behavior. If the noise level is too high, the policy may handle terrain by making contact rather than relying on perception. Conversely, if the noise level is too low, the sim-to-real gap increases, leading to potential deployment failures in the real world. To balance vision and proprioception, we introduce DAgger to make the student policy learn teacher actions generated from precise perception. Some DAgger-based teacher-student frameworks, such as \cite{cheng2023extreme}, utilize supervised learning to force the student to output the same actions as the teacher. Since the state transition is determined by the current state of the student policy and the output action, a gap arises between the teacher and the student, reducing the teacher's effectiveness. To address this, we propose a novel teacher-student framework that uses reinforcement learning to continue exploring the space based on the teacher’s supervision signal. The policy loss is a weighted sum of the Mean Squared Error between the actions from the teacher policy $\pi_{teacher}$ and the student policy $\pi_{student}$, and the loss of PPO (clip version)\cite{schulman2017proximal}:

\begin{equation}
 \mathcal{L}_{\text{distillation}} = \left\| \pi_{\text{teacher}}(s_t) - \pi_{\text{student}}(s_t) \right\|^2_2 
\end{equation}

\begin{align}
L^{\text{PPO}}(\theta) = & \mathbb{E}_t \left[ \min \left( r_t(\theta) \hat{A}_t, \text{clip}(r_t(\theta), 1 - \epsilon, 1 + \epsilon) \hat{A}_t \right) \right] \nonumber \\
& - c_1 \mathbb{E}_t \left[ (V_\theta(s_t) - R_t)^2 \right] + c_2 \mathbb{E}_t \left[ S[\pi_\theta](s_t) \right]
\end{align}

\noindent where:
\begin{itemize}
    \item \( r_t(\theta) = \frac{\pi_\theta(a_t | s_t)}{\pi_{\theta_{\text{old}}}(a_t | s_t)} \) is the probability ratio between the new and old policies.
    \item \(\hat{A}_t\) is the estimated advantage function.
    \item \(\epsilon\) is the clipping parameter.
    \item \(V_\theta(s_t)\) is the estimated value function.
    \item \(R_t\) is the return.
    \item \(S[\pi_\theta](s_t)\) is the entropy of the policy, which encourages exploration.
    \item \(c_1\) and \(c_2\) are coefficients that balance the value function loss and the entropy bonus, respectively.
\end{itemize}

The policy loss term ensures that the new policy does not deviate too much from the old policy by clipping the probability ratio. The value function loss term minimizes the difference between the estimated value and the actual return. The entropy bonus term encourages the policy to explore more by maximizing the entropy of the policy distribution.
This loss function helps PPO achieve stable and efficient training by balancing exploitation and exploration. Our overall loss function can be expressed as:

\begin{equation}
 \mathcal{L}_{\text{total}} = \alpha\mathcal{L}_{\text{distillation}} + \beta\mathcal{L}_{\text{PPO}}
 \label{equation_loss}
\end{equation}

$\alpha$ and $\beta$ are coefficients that balance the different losses. Our Distillation-PPO method trains both the policy and the history state encoder using gradients from DAgger and PPO. Compared to approaches based solely on PPO or DAgger, which only resume weights, our method not only enables the policy to fine-tune according to performance but also allows it to learn from the teacher. We present the schematic diagram of the D-PPO training framework in Fig.~\ref{fig:main}.

% \subsection{Rewards on Humanoid Robots Perceptive Locomotion}
\subsection{Rewards on Humanoid Robots Perceptive Locomotion}
To enable humanoid robots to navigate complex and unknown environments, it is essential to use sensors for environmental perception. In our perception framework, we utilize the approach from~\cite{miki2022elevation} to construct an elevation map. Compared to quadruped robots, humanoid locomotion is inherently less stable, making pose estimation more challenging. We implement a LiDAR-Inertial Odometry~(LIO) system on our humanoid robots to achieve accurate pose and mapping, following the method in~\cite{xu2022fast}. In addition, the compact structure of humanoids often leads to depth map occlusions caused by the legs. To mitigate this, we incorporate joint angles $q$ and a forward kinematic model to estimate the robot's body bounding box within the depth image, allowing us to exclude the occluded regions. Subsequently, we input the point cloud and pose into the elevation map framework and update the height of each grid using a one-dimensional Kalman filter as follows:
\begin{equation}
    h = \frac{\sigma_p^2 h + \sigma_m^2 p_z}{\sigma_m^2 + \sigma_p^2}
    \label{equation_vision}
\end{equation}
where \( \sigma_p^2 \) represents the variance derived from the sensor noise model, and \( p_z \) is the height measurement. The term \( \sigma_m^2 \) is the estimated variance of the cell. The noise model is given by \( \sigma_p^2 = \alpha_d d^2 \), with \( d \) as the distance from the sensor and \( \alpha_d \) as a hyperparameters. The variance of each grid is updated according to,
\begin{equation}
    \sigma_m^2 = \frac{\sigma_m^2 \sigma_p^2}{\sigma_m^2 + \sigma_p^2}
\end{equation}
During the student policy training, we add the Gaussian noise to each grid to bridge the gap between simulation and real sensors.

To achieve a stable locomotion gait, we build our comprehensive reward system with periodic, tracking and regularized rewards. Following the approach in~\cite{siekmann2021sim}, we define the periodic reward based on the swing and stance phases of bipedal locomotion. The swing phase, where the foot moves through the air, is followed by the stance phase, where the foot is planted on the ground. Each reward component is defined by a coefficient $\alpha_i$, a phase indicator $I_i(\phi)$, and a phase-specific reward function $V_i(s_t)$. $\phi$ represents the cycle time, and $i$ denotes whether the phase is a swing or stance phase. The phases are sequentially arranged to cover the entire cycle, with the swing phase lasting a proportion $\rho \in (0,1)$ and the stance phase occupying $1-\rho$. The reward for a single foot is formualted as follows:
\begin{equation}
\begin{aligned}
&r_{per} = \sum \alpha_i \mathbb{E}[I_i(\phi)] V_i(s_t)\\
&V_{stance}(s_t) = \text{exp}(\lambda_1F_{f}^2)\\
&V_{swing}(s_t) =  \text{exp}(\lambda_2v_{f}^2)
\end{aligned}
\end{equation}
where $F_{f}$ is the norm force of each foot, $v_{f}$ is the speed of each foot. For modeling the phase indicator $I_i(\phi)$, we follow the \cite{park2001impedance} to use the mathematical expectation of Von Mises distribution. And we formulate as, 
\begin{equation}
\begin{aligned}
    &Q_1=I_{stance}(\phi+\theta_{left}) \\
    &Q_2=I_{stance}(\phi+\theta_{right}) 
\end{aligned}
\end{equation}
where the $\theta_{left}$, $\theta_{right}$ is the offsets of left and right leg in cycle time.

To strengthen the robustness of sim-to-real transfer, we integrated regularization rewards into the overall reward framework. These rewards impose motion constraints that prioritize both smoothness and safety. As detailed in Table~\ref{tab: regularization rewards}, the DoF limits reward ensures that movements stay within the robot's physical capabilities, preventing actions that could damage its mechanisms or exceed its operational scope. $\mathbf{b}$ represents the joint angle, with $\mathbf{b}_{lower}$ and $\mathbf{b}_{upper}$ being the lower and upper bounds, respectively. The DoF velocity reward maintains optimal movement speeds, while the DoF acceleration reward promotes smooth acceleration and deceleration, enhancing overall stability. For robotic arms, an arm DoF penalty discourages excessive movements. Additionally, rewards related to torso yaw and orientation differentials are crucial for maintaining balance and proper orientation, especially in humanoid robots.

The command reward encourages the robot to move alone in the specific directions formulated as:
\begin{equation}
\vspace{-1mm}
r_{com}=\sum\lambda_i\text{exp}(-\omega_i(|v^{i}_{des}-v^{i}_{t}|))\quad i \in(x,y,yaw)
% \vspace{-0.7mm}
\end{equation}
where $\lambda_i$ and $\omega_i$ are the weight of each direction of command rewards, $\mathbf{v}_{des}$ is the desire velocity along the specific direction and $\mathbf{v}_t$ is the velocity at $t$.
\begin{table}[]
    \centering
    \caption{Regularization Rewards}
    \scalebox{1}{
\begin{tabular}{ll}
\hline
Item                     & Detail \\ \hline
Action differential      & $\text{exp}(\alpha_i\|\mathbf{a}_t - \mathbf{a}_{t-1}\|_2)$   \\
DoF limits               & $\text{exp}(\alpha_i(\text{min}(0,\mathbf{b} - \mathbf{b}_{upper})$\\
                         &$-\text{max}(0,\mathbf{b}-\mathbf{b}_{lower}))) $  \\
DoF velocity             & $\text{exp}(\alpha_i\|\dot{\mathbf{b}}\|^2_2)$   \\
DoF acceleration         & $\text{exp}(-\alpha_i\|\ddot{\mathbf{b}}\|^2_2)$   \\
Arm DoF penalty          & $\text{exp}(\|\mathbf{b}_{arm}\|_1)$   \\
Orientation differential & $\text{exp}(\alpha_i(roll^2+pitch^2))$   \\
Torso yaw                & $\|b_{torso~yaw}\|_1$   \\ 
Torques                  & $\text{exp}(\alpha_i\|\mathbf{\tau}_t\|_2)$    \\\hline
\end{tabular}
    }
    \vspace{-2mm}
    \label{tab: regularization rewards}
\end{table}

%%%%%%%%%%%%%%%%%%%%%%%%%%%%%%%%%%%%%%%%%%%%%%[section]%%%%%%%%%%%%%%%%%%%%%%%%%%%%%%%%%%%%%%%%%%%%%%%%%%%%%%%%%%
\section{Experiments}

\subsection{Training and Implementation Details}
We train both teacher and student policy on a single 4090 GPU on 4096 Isaac Gym\cite{makoviychuk2021isaac} environments in parallel. In our perception system, we use Livox MID-360 and Xsence Inertial measurement unit~(IMU) to get pose by Fast-LIO2~\cite{xu2022fast}.  We utilize an Orbbec 355L depth camera to obtain the depth image and convert it into point cloud in the global frame.

\begin{figure}[t]
	\centering
	\includegraphics[width=\linewidth]{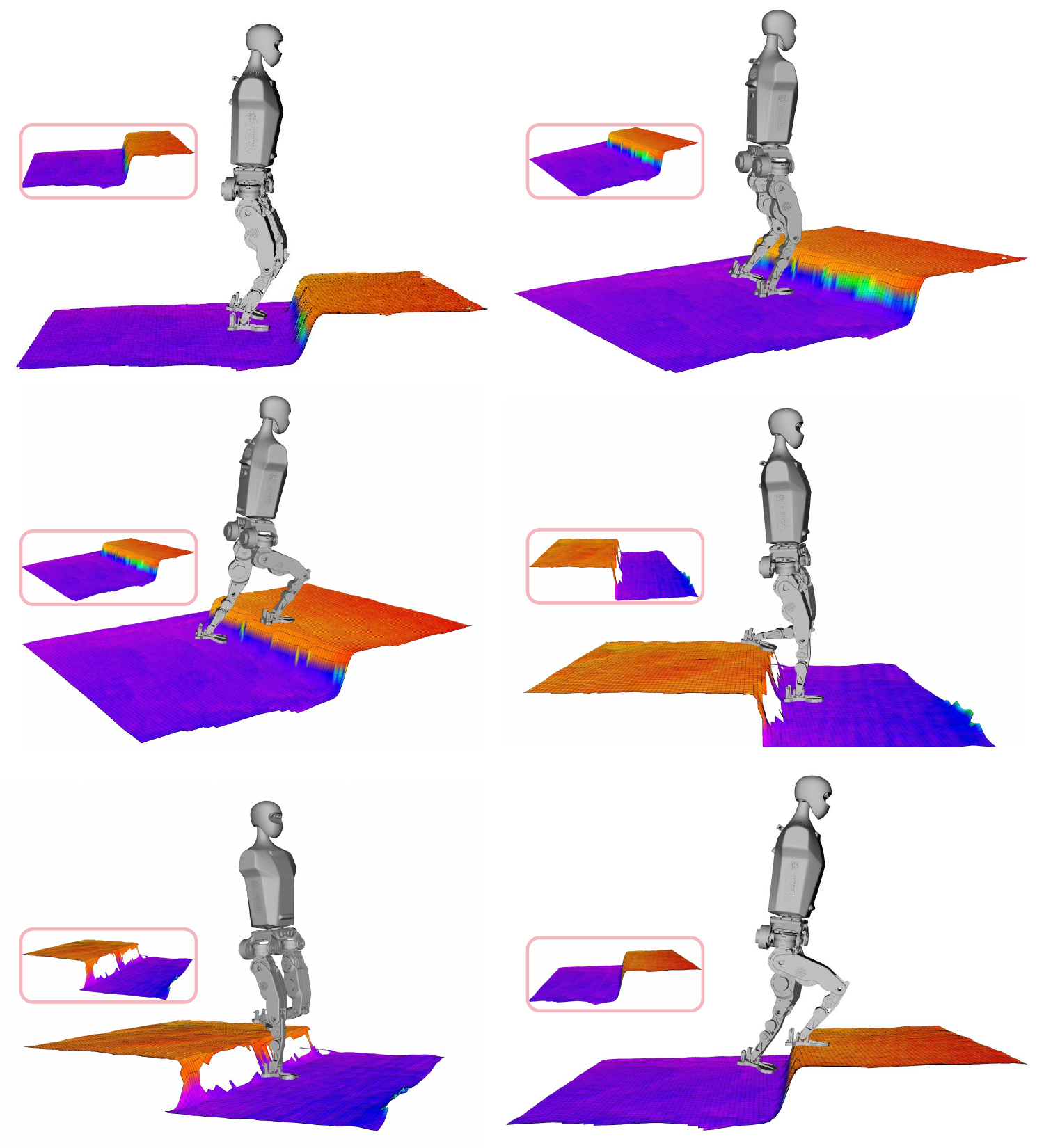}
 \caption{We provide a detailed demonstration of the visual perception system's performance in estimating posture on elevated terrains and during motion. The figure shows that our system can accurately estimate and reconstruct the terrain. By selecting scan points on this high-precision reconstructed terrain and leveraging the noise robustness trained by D-PPO, our robot can adapt well to various complex terrains.}
	\label{fig:terrain}
	% \vspace{-0.55cm}
\end{figure}

In humanoid locomotion tasks, the actions are defined by $a_t \in \mathbb{R}^{18}$, which include the desired positions for each actuated joint. Observations contain the current linear and angular velocities, the average velocities in $(x, y, \text{yaw})$, and the orientation of the pelvis relative to the local frame, as well as the position and velocity of each joint in the legs and arms, and the action from the previous time step. Beyond proprioceptive data, the command $c_t = (v^{x}, v^{y}, \omega^{\text{yaw}})$ is used to guide the humanoid's movement according to the desired trajectory. Additionally, periodic signals, including the sine and cosine of the cycle time the swing phase ratio $\rho$, and the scan dots of terrain geometry are incorporated.
We followed the all training parameter settings from legged gym~\cite{rudin2022learning}. For the overall noise signal, we chose the same settings as ~\cite{zhang2024whole}, as we believe this noise level accurately reflects the real-world interactions of the robot. To balance the two learning losses, we set $\alpha = 0.5$ and $\beta = 0.5$ in equation~\ref{equation_loss}.

\subsection{Perceptive Locomotion Performance}

\begin{figure}[t]
	\centering
	\includegraphics[width=\linewidth]{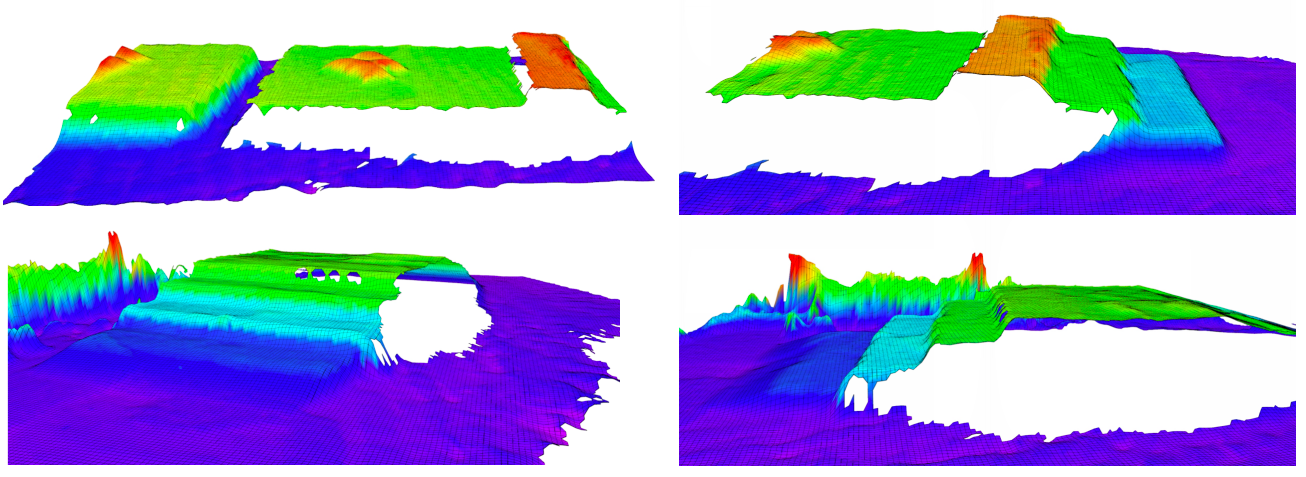}
 \caption{High precision demonstration in terrain reconstruction.}
	\label{fig:visualization}
	% \vspace{-0.55cm}
\end{figure}

\begin{figure}[t]
	\centering
	\includegraphics[width=\linewidth]{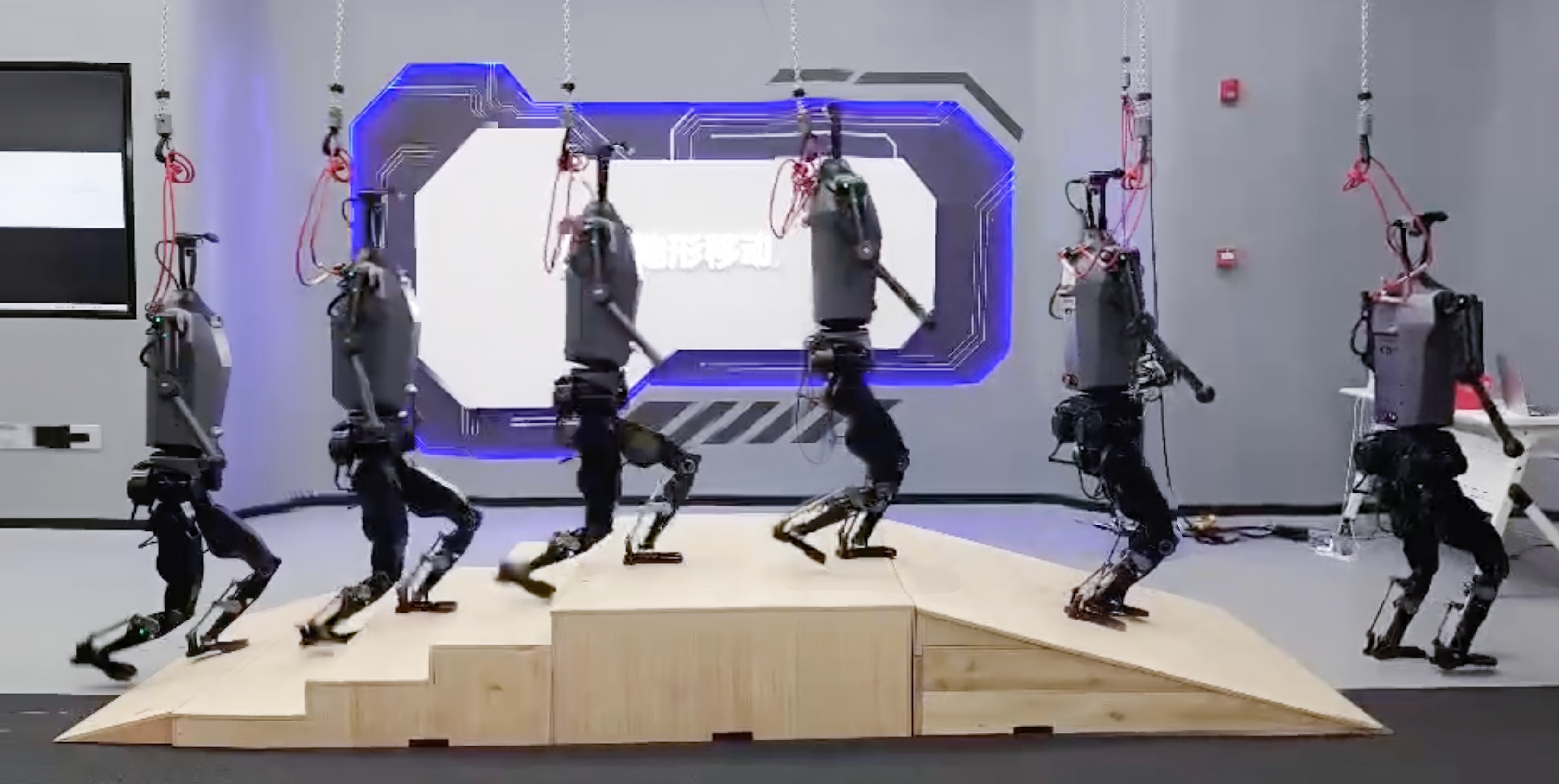}
 \caption{The humanoid robot Tien Kung accurately walks on stair terrain and can descend slopes with ease.}
	\label{fig:stairs}
	% \vspace{-0.55cm}
\end{figure}

We present the visualization results of the terrain and the robot's posture throughout the entire process of navigating a high platform in Fig.~\ref{fig:terrain}. Specifically, we details each step of the robot's movement on the high platform terrain, including starting, climbing, and descending. These visualizations clearly show the changes in the robot's posture and adjustments in its actions at different stages.

We show the terrain reconstruction results for various terrains during the actual operation of our robot in Fig.~\ref{fig:visualization}. Thanks to our LIO module, we can achieve highly accurate terrain reconstruction, allowing us to select very precise scan points and obtain more accurate terrain information. Fig.~\ref{fig:visualization} illustrates the reconstruction results on different terrains, including flat ground, slopes, steps, and irregular terrains. These results demonstrate how the robot utilizes the perception system to adapt to different terrain changes.

In Fig.~\ref{fig:stairs}, we display the process of the Tiangong humanoid robot accurately completing tasks on classic stair and slope terrains after being trained with D-PPO. Fig.~\ref{fig:stairs} details the robot's walking process on stair and slope terrains, including ascending stairs, descending stairs, ascending slopes, and descending slopes. The distinct differences in the robot's posture on different terrains are clearly visible. For example, when ascending stairs, the robot raises its knees to step over each step, while when descending slopes, the robot adjusts its center of gravity to maintain balance. These results indicate that the robot's adaptability and stability on complex terrains have been significantly improved after training with D-PPO.

\begin{table}[h]
\centering
\begin{tabular}{|c|c|c|c|}
\hline
\textbf{} & \textbf{Training} & \textbf{Control} & \textbf{Noise} \\
\textbf{} & \textbf{Difficulty} & \textbf{Performance} & \textbf{Robustness} \\
\hline
Only Distillation & Easy & Normal & Worse \\
\hline
Only RL & Hard & Normal & Medium \\
\hline
D-PPO(Ours) & \textcolor{violet}{\textbf{Easy}} & \textcolor{violet}{\textbf{Good}} & \textcolor{violet}{\textbf{Good}} \\
\hline
\end{tabular}
\caption{Comparison of Different Methods}
\label{table:comparison}
\end{table}

As shown in Table \ref{table:comparison}, we illustrate the comparison of different methods in terms of training difficulty, control performance, and noise robustness. It can be observed that traditional methods relying solely on distillation, while having lower training difficulty, perform poorly in noise robustness. On the other hand, methods relying solely on reinforcement learning show improved noise robustness but have higher training difficulty. In contrast, our proposed D-PPO method excels in all aspects. Specifically, D-PPO not only maintains low training difficulty but also significantly enhances control performance and noise robustness. That demonstrates D-PPO method effectively combines the advantages of distillation and reinforcement learning, providing an efficient and stable training framework suitable for humanoid robot locomotion control in complex terrain environments.

When visual information is incorporated, we face more complex and sensitive data. Directly using distillation makes it difficult to cover all states. When encountering corner cases or errors from the teacher, the student struggles to learn effectively. Besides, learning the vision-action mapping through RL is very challenging, as the current data sampling efficiency of RL cannot ensure the learning of such large-scale visual information. D-PPO combines distillation and RL to ensure both training stability and learning efficiency. Our method can be applied not only to humanoid robots but also to any other robots requiring visual perception.

% %%%%%%%%%%%%%%%%%%%%%%%%%%%%%%%%%%%%%%%%%%%%%%[section]%%%%%%%%%%%%%%%%%%%%%%%%%%%%%%%%%%%%%%%%%%%%%%%%%%%%%%%%%%
% \section{Conclusion}
% \label{sec:conclusion}

%%%%%%%%%%%%%%%%%%%%%%%%%%%%%%%%%%%%%%%%%%%%%%[section]%%%%%%%%%%%%%%%%%%%%%%%%%%%%%%%%%%%%%%%%%%%%%%%%%%%%%%%%%%

\bibliographystyle{IEEEtran}
\typeout{}
\bibliography{IEEEabrv,mybibfiles}
% \theendnotes
\end{document}